\documentclass[12pt]{article}

\usepackage{arxiv}
\usepackage{caption}
\usepackage[utf8]{inputenc} 
\usepackage[T1]{fontenc}    
\usepackage{hyperref}       
\usepackage{url}            
\usepackage{booktabs}       
\usepackage{threeparttable} 
\usepackage{amsfonts}       
\usepackage{nicefrac}       
\usepackage{microtype}      
\usepackage{graphicx}
\usepackage[numbers]{natbib}
\usepackage{doi}
\usepackage{subcaption} 
\usepackage{float}      
\usepackage{tikz}
\usetikzlibrary{arrows.meta,positioning,calc,shapes.geometric,shapes.symbols,shapes.misc,fit,backgrounds}
\usepackage{amsmath}
\tikzset{>=Latex}
\usepackage{rotating}
\usepackage{placeins}

\title{Fine-Tuning Open-Weight Language Models to Deliver Cognitive Behavioral Therapy for Depression: A Feasibility Study}

\author{
	Talha Tahir \\
	Department of Psychiatry\\
	University of Toronto\\
	\texttt{talhat.tahir@mail.utoronto.ca} \\
}

\date{}

\renewcommand{\headeright}{}
\renewcommand{\undertitle}{}
\renewcommand{\shorttitle}{Fine-tuning LLMs for CBT}

\hypersetup{
hidelinks=true,
pdftitle={Fine-Tuning Open-Weight Language Models to Deliver Cognitive Behavioral Therapy for Depression: A Feasibility Study},
pdfsubject={cs.AI, cs.CL},
pdfauthor={Talha Tahir},
pdfkeywords={Large Language Models, Cognitive Behavioral Therapy, Depression, Fine-tuning, Open-weight models, Synthetic data, AI Therapist},
}

\begin{document}
\maketitle

\begin{abstract}
\noindent
Cognitive Behavioral Therapy (CBT) is a well-established, evidence-based treatment for Major Depressive Disorder. Unfortunately, there exist significant barriers to individuals accessing CBT, including cost, scarcity of therapists and stigma. This study explores the feasibility of fine-tuning small open weight large language models (LLMs) to deliver CBT for depression. Using synthetic CBT transcripts generated by the Nous Research fine-tune of Llama 3.1 405b, we fine-tuned three models: Mistral 7b v0.3, Qwen 2.5 7b, and Llama 3.1 8b. CBT fidelity was evaluated through a modified Cognitive Therapy Rating Scale (CTRS). All fine-tuned models were compared against each other, as well as their instruct-tuned variants. Simulated patient transcripts were generated for the purpose of evaluating model performance, with the instruct and CBT-tuned models acting as the therapist and DeepSeek-V2.5 acting as the patient. These simulated transcripts were evaluated on a modified CTRS by Gemini 1.5 Pro-002. Our findings demonstrated that the CBT-tuned models significantly outperformed their instruct-tuned counterparts, with an average improvement of 11.33 points (p < 0.001) on total CTRS score. Llama 3.1 8b had the strongest performance (mean CTRS score 67.86 ± 7.24), followed by Qwen 2.5 7b (64.28 ± 9.55) and Mistral 7b v0.3 (64.17 ± 9.79), with these differences between models being statistically significant. The CBT-tuned models were competent in implementing core CBT techniques and providing empathetic responses, however, there were limitations observed in agenda adherence, exploration depth and long-context coherence. This study establishes that CBT specific fine-tuning can effectively encode therapeutic competencies in small LLMs, though significant technical and ethical considerations must be resolved prior to clinical deployment.
\end{abstract}

\section{Introduction}
Major depressive disorder (MDD) is a common psychiatric condition that affects 20.6\% of Americans over their lifetime \cite{Hasin2018EpidemiologyStates}. Individuals with MDD experience significant decrements in social functioning, role-emotional functioning, and cognition, resulting in substantial economic burden \cite{Hasin2018EpidemiologyStates,Greenberg2021The2018}. The yearly economic cost of MDD in 2018 was calculated to be \$326.2 billion, significantly increased from \$236.6 billion in 2010 \cite{Greenberg2021The2018}. Despite, the functional impairment and economic loss associated with MDD, many Americans are not able to access evidence-based treatments \cite{Gonzalez2010DepressionFew}. 

Cognitive behavioral therapy (CBT) is one of the most effective non-pharmacologic interventions for MDD, demonstrating comparable efficacy to antidepressants \cite{Hundt2013TheLiterature, Cuijpers2013ATreatments,Cuijpers2019EffectivenessMeta-analysis,Beck1970CognitiveTherapy,DeRubeis1999MedicationsComparisons}. CBT is a time limited manualized therapy that focuses on identifying and modifying maladaptive thought patterns and behaviors that contribute to symptoms of depression \cite{Beck1970CognitiveTherapy}. Despite the well-established efficacy of CBT in individuals with MDD, it remains underutilized due to a combination of patient, provider and systemic factors. Notable barriers include, fear of stigma, cost, insufficient number of trained therapists and poor access to mental health care in certain geographic regions \cite{Collins2004GapsCare}. 

Many of the aforementioned challenges could be addressed through the deployment of a CBT delivery system that is powered by artificial intelligence (AI). AI therapists have the potential to offer personalized, scalable and cost-effective psychotherapy to individuals who might otherwise struggle to access face-to-face CBT \cite{Obradovich2024OpportunitiesPsychiatry}. The rapid advancement of large language models has perhaps, for the first time, created an opportunity to craft AI therapists that can deliver true manualized CBT \cite{Obradovich2024OpportunitiesPsychiatry, Stade2024LargeEvaluation}. LLMs excel at text completion, question answering and language understanding \cite{Lappin2024AssessingModels}. Their ability to produce context aware responses and mimic the style of any given template, positions them well to deliver a manualized therapy like CBT \cite{Stade2024LargeEvaluation, Jiang2024ATherapyb}. 

There have been several papers published within the past two years that have attempted to use LLMs to either augment CBT or provide CBT directly to patients \cite{Shen2024AreTherapy}. Some have used prompt engineering approaches, where an existing closed source model is provided with a specialized prompt to steer its responses to be informed by CBT principles \cite{Iftikhar2024TherapyCBT, Kian2024CanStudents}. HELPERT is one such example, that utilized GPT-4 with a prompt that was designed to guide users through self-reflection and problem solving \cite{Iftikhar2024TherapyCBT}. Other approaches have targeted only a small aspect of the CBT problem space, such as challenging cognitive distortions or Socratic questioning \cite{Izumi2024ResponseQuestioning, Na2024CBT-LLM:Answering}. 

Unfortunately, there are significant limitations to the previously discussed approaches. Prompt engineering for example, can be brittle and is limited by the context of the model \cite{Knoth2024AIStrategies}. There are often detrimental effects on model coherence when passing too large of a system prompt \cite{Levy2024SameModels}. Furthermore, given that CBT is a manualized therapy where the dialogue and content can vary based on the phase, it would be necessary to engineer multiple prompts for different sessions and situations \cite{Beck1970CognitiveTherapy}. Even with situation specific prompts, it would be difficult to provide a model with all of the information that is required to deliver CBT. Large frontier models such as Claude 3.5 Sonnet, GPT-4o and Llama 3.1 405b would likely not suffer from a lack of knowledge about CBT and may perform well with only prompt engineering. However, most large frontier models are closed source and require significant compute infrastructure for inference \cite{Hoffmann2022TrainingModels}. To our knowledge, there are no papers that have fine-tuned an open weight LLM to deliver a full course of CBT for depression. 

In this paper, we propose a method for fine-tuning three small versatile LLMs, Mistral 7b v0.3, Qwen 2.5 7b and Llama 3.1 8b to deliver CBT for depression \cite{Dubey2024TheModels, QwenTeam2024Qwen2.5:Models, Yang2024Qwen2Report, Jiang2023Mistral7B}. Although such a task would be considerably easier if we utilized larger open-weight LLMs, it would run counter to the goal of our approach, which is to contribute to the development of an AI therapist that can deliver cost-effective, accessible and personalized care to individuals with MDD. The training and inference costs associated with highly parameterized models would also exceed our compute budget. 

\section{Materials and methods}
\subsection{Overview}
This study aims to fine-tune small open-weight large language models on synthetically generated CBT transcripts to perform CBT for depression. We start by creating a diverse set of CBT transcripts using the Nous Research fine-tune of Llama 3.1 405b \cite{Teknium2024HermesReport, Dubey2024TheModels}. We fine-tune Mistral 7b v0.3, Qwen 2.5 7b and Llama 3.1 8b on the synthetic transcripts \cite{Dubey2024TheModels, QwenTeam2024Qwen2.5:Models, Yang2024Qwen2Report, Jiang2023Mistral7B}. We then run simulated patient interactions between our fine-tuned models, who act as the therapist, and DeepSeek-V2.5, which acts as the patient \cite{I.2024DeepSeek-V2:Model}. The transcripts that are generated through these simulated therapy sessions are then evaluated on a modified Cognitive Therapy Rating Scale using Google's Gemini 1.5 Pro-002 \cite{Goldberg2020TheScale, Affrunti2019TheClinicians, Team2024GeminiContext}.

\subsection{Synthetic data generation}
A total of 58 sets of CBT transcripts were generated using the Nous Research fine-tune of Llama 3.1 405b \cite{Teknium2024HermesReport, Dubey2024TheModels}, with this number representing the maximum feasible quantity given our compute budget. Each set consisted of 20 sessions and was meant to represent a therapy course for one unique patient suffering from MDD. Patient profiles were generated through random selection of characteristics such as age, gender, ethnicity, education level, occupation, symptom severity, engagement level, life events, family background, hobbies, social support systems, personality traits, and coping mechanisms. 

As each set was meant to represent a full course of CBT for depression for a particular patient, continuity between sessions in a set was essential. To create these residual connections between sessions, each session concluded with a concise summary detailing the key points discussed, techniques used, and homework assigned. This summary was then paraphrased in the patient’s first message of the next session to allow for growth and progression in the narrative. 

Another important aspect of generating synthetic data was to employ the appropriate core CBT techniques across different phases of the therapy \cite{Kazantzis2018TheMeta-Analyses, Fenn2013TheTherapy}. To this end, we divided up the sessions within a course of CBT into four distinct phases: assessment (sessions 1-3), initial (sessions 4-7), middle (sessions 8-17), and termination (sessions 18-20). Each phase was associated with its own goals and CBT techniques. The assessment phase focused on gathering detailed information about the patient's depression symptoms, life circumstances, and treatment history while beginning to establish the therapeutic relationship. The initial phase aimed to introduce the cognitive conceptualization framework and begin socializing the patient to CBT principles and techniques. The middle phase built upon these principles with deeper exploration and modification of cognitive distortions and maladaptive beliefs. Thought records, behavioral experiments and cognitive restructuring were the main foci of this phase. The termination phase involved consolidating gains from therapy, planning for future challenges and exploring relapse prevention. Our division of CBT phases and associated techniques was informed by the CBT literature \cite{Kazantzis2018TheMeta-Analyses, Fenn2013TheTherapy, McGinn2000CognitiveStatus, Hundt2013TheLiterature, Fennell1987CognitiveChange, Carter2015PredictorsProcess}.

Each session within a set was generated in four parts. The first part consisted of a brief check-in with the patient and setting an agenda for the session. The second and third parts focused on discussing previous homework, exploring new techniques relevant to the session phase or building upon techniques introduced in previous sessions. A session concluded with a review of key points, a summary provided to the patient, and homework assigned for the next meeting \cite{Kazantzis2018TheMeta-Analyses}.

Built into the synthetic transcripts, was an acknowledgment of the limitations of an AI therapist. In every transcript, the AI therapist informed the patient about the possibility of hallucinations, inability to read non-verbal cues and requested that the patient seek emergency assistance if experiencing ongoing thoughts of self-harm or suicide. 

\subsection{Model selection and fine-tuning}
The three open-weight models that were selected for fine-tuning were Mistral 7b v0.3, Qwen 2.5 7b, and Llama 3.1 8b. These models were chosen for their robust performance compared to other models of their size, long context windows and wide community support for fine-tuning \cite{Dubey2024TheModels, QwenTeam2024Qwen2.5:Models, Yang2024Qwen2Report, Jiang2023Mistral7B}. We chose to use models in the 7-8 billion parameter range to balance inference costs with model capabilities. We postulated that models smaller than 7 billion parameters would not have the capacity to be fine-tuned for a complex task like CBT, and models in the 20-70 billion parameter range would be outside of our compute budget.

Due to computational constraints, we opted for the Quantized Low-Ranked Adaptation (QLoRA) fine-tuning technique using Unsloth \cite{Dettmers2023QLoRA:LLMs, unslothai2024Unsloth}. We chose the Unsloth library for its built-in memory optimizations which allowed us to fine-tune each model on a single NVIDIA A40 graphics card \cite{unslothai2024Unsloth}. The 4-bit quantized base versions of the models were fine-tuned for one epoch over the training set. To enhance inference fidelity, the fine-tuned adapters were merged with 16-bit versions of the respective base models. 

\subsection{Simulated CBT sessions}
The fine-tuned and instruct-tuned versions of Mistral 7b v0.3, Qwen 2.5 7b  and Llama 3.1 8b were employed as AI therapists in simulated CBT sessions, with DeepSeek-V2.5 acting as the patient \cite{Dubey2024TheModels, QwenTeam2024Qwen2.5:Models, Yang2024Qwen2Report, Jiang2023Mistral7B, I.2024DeepSeek-V2:Model}. We utilized the instruct-tuned versions of the base models as a point of comparison. For both the CBT fine-tuned and instruct-tuned models, inference was performed with llama.cpp using 8-bit quantized GGUFs \cite{ggerganov2024Llama.cpp}.

Similar to the synthetic transcript generation process, the simulated CBT sessions were generated in 5 sets of 20 sessions. The same 5 distinct patient profiles were used for simulations with each model variant, to ensure consistency for statistical comparisons. For every session, the simulation was run until the AI therapist generated a session summary, if 50 turns of conversation were completed or if the transcript exceeded 5000 words. After the termination of a session, a session summary was generated through a separate prompt to claude-3.5-sonnet. For both the CBT and instruct-tuned model simulations, the session summaries from one session were presented to the AI therapist in the patient’s first message of the next session. This ensured continuity across sessions. There were no differences in prompting the CBT-tuned or instruct-tuned models other than using the prompt templates specific to each model. 

\subsection{Evaluation framework}
The transcripts from the simulated CBT sessions were graded through an automated evaluation pipeline. Specifically, Gemini 1.5 Pro-002 was used to evaluate the transcripts according to a modified Cognitive Therapy Rating Scale (CTRS). The CTRS is the gold standard for assessing CBT fidelity and involves rating a therapy transcript or recording on the following 11 categories: Agenda, Feedback, Understanding, Interpersonal Effectiveness, Collaboration, Pacing and Efficient Use of Time, Guided Discovery, Focusing on Key Cognitions or Behaviors, Strategy for Change, Application of Cognitive-Behavioral Techniques and Homework \cite{Goldberg2020TheScale, Affrunti2019TheClinicians}. 

Although the CTRS has only been validated with human therapists as raters, there is a precedent to deploying it with AI systems \cite{Flemotomos2021AutomatedRepresentations}. A 2021 paper used a BERT-based model for automated scoring on the CTRS with encouraging results \cite{Flemotomos2021AutomatedRepresentations}. Our decision to utilize AI raters rather than human raters was informed by multiple factors including rater consistency, scalability and the advancing capabilities of frontier models. Gemini-1.5-Pro-002 was chosen for its excellent recall on long context tasks, state of the art reasoning abilities and free availability through Google’s Gemini API \cite{Team2024GeminiContext}. Furthermore, frontier LLMs such as Gemini-1.5-Pro-002 have been trained on trillions of tokens, resulting in a robust knowledge base in a variety of fields, including CBT \cite{Ke2024ExploringReview}. 

To increase the resolution of the ratings, the CTRS scale was expanded from a 0-6 Likert scale to a 1-10 scale. Furthermore, given the issues LLMs have experienced when processing numbers and arithmetic, we opted to convert the numeric Likert scale to word-based ratings that ranged from ‘Severely Deficient’ to ‘Outstanding’ \cite{Feng2024HowLLMs}. We acknowledge that the aforementioned modifications to the CTRS will affect its validity in unpredictable ways. However, for the purposes of this pilot study, with minimal human input, we felt that it was appropriate to use the CTRS as a scaffold for our automated evaluation process. For each transcript, Gemini-1.5-Pro-002 was asked to provide both the word-based ratings for each CTRS category as well as justifications for that rating with specific examples. In cases where category evaluation failed after multiple retries, the system employed a conservative approach, assigning the lowest possible score and documenting the failure in the justification. The word-based ratings were then converted back to an integer between 1 and 10 for the statistical analysis. 

For each transcript, the patient’s first and the therapists last statement were removed prior to passing the transcript to Gemini-1.5-Pro-002 for evaluation. This was done to avoid sending the summary of the previous or current session to the evaluation model, as those summaries, which contained many references to CBT techniques, could have biased the evaluations.  

\subsection{Analysis}
Our analysis employed a mixed-effects model, to account for both fixed effects (model type, variant, session number) and random effects (patient-specific variations) \cite{Oberg2007LinearModels}. This approach was chosen to address the nested nature of the data, where multiple sessions were nested within patients, and to account for the non-independence of observations within patient groups. The primary analysis used a linear mixed-effects model with the total CTRS score as the dependent variable. The model was specified as follows:

\begin{equation}
\text{total\_score} \sim \mathcal{C}(\text{model}) + \mathcal{C}(\text{variant}) + \text{session\_centered} + (1 \mid \text{patient\_id})
\end{equation}

In the formula above, session\_centered represents the z-standardized session number, model and variant are categorical predictors, and (1 | patient\_id) denotes random intercepts for each patient. Fixed effects included the main effects of model type, variant, and session number, without interaction terms. The model was fitted with a maximum likelihood estimate using the LBFGS optimizer \cite{Liu1989OnOptimization}. When convergence issues were encountered, the Powell optimizer was used \cite{Steihaug2013GlobalFunction}. 

We implemented the Bonferroni correction method for multiple comparisons of fixed effects \cite{Napierala2012WhatCorrection}. We validated our model by assessing residual normality through Q-Q plots and homoscedasticity using residual versus fitted value plots \cite{Schutzenmeister2012CheckingPlots}. Model fit was assessed using both Akaike Information Criterion (AIC) and Bayesian Information Criterion (BIC) \cite{Kuha2004AICPerformance}. The explained variance was calculated as the ratio between fitted values variance and total score variance.

Prior to our analysis, we employed checks for missing values, verified group balance, and assessed predictors for co-linearity. We excluded all cases with missing values in key variables such as total score, model, variant, session and patient ID.  

\section{Results}
\subsection{Simulated patient transcripts}
For each instruct-tuned model and its CBT-tuned variant, 100 transcripts across five simulated patients were generated. Once the transcripts from the simulated patient interactions were cleaned, via removal of the patient’s first and therapist’s last statements, a total of 593 transcripts remained. All seven of the removed transcripts belonged to Mistral 7b v0.3 instruct, and were excluded as there was no text left to analyse after the cleaning process above was completed. Mistral 7b v0.3 instruct had initiated an early session termination in all of the excluded transcripts. 

\subsection{Overall performance and model comparison}
The CBT-tuned variants of the three base models, significantly outperformed the instruct-tuned versions. See Figure 1 for an overview of model performance and score distribution by CTRS domain. Llama 3.1 8b was the strongest performer in terms of mean CTRS score (67.86 ± 7.24), followed by fine-tuned Qwen 2.5 7b (64.28 ± 9.55) and fine-tuned Mistral 7b v0.3 (64.17 ± 9.79). In the instruct-tuned category, Llama 3.1 8b was again the best performing model (59.83 ± 10.44), beating Qwen 2.5 7b (55.09 ± 9.91) and Mistral 7b v0.3 (47.20 ± 14.09).

Fine-tuning on the synthetic transcript dataset produced statistically significant improvement in CTRS scores for all models. Mistral 7b v0.3 benefitted most from fine-tuning with a 16.97-point increase in mean CTRS score in comparison to the instruct-tuned version. However, both Llama 3.1 8b and Qwen 2.5 7b also demonstrated significant improvements of 8.03 and 9.19 points, respectively. \newline

\begin{figure}[h!]
    \centering
    \includegraphics[width=\textwidth]{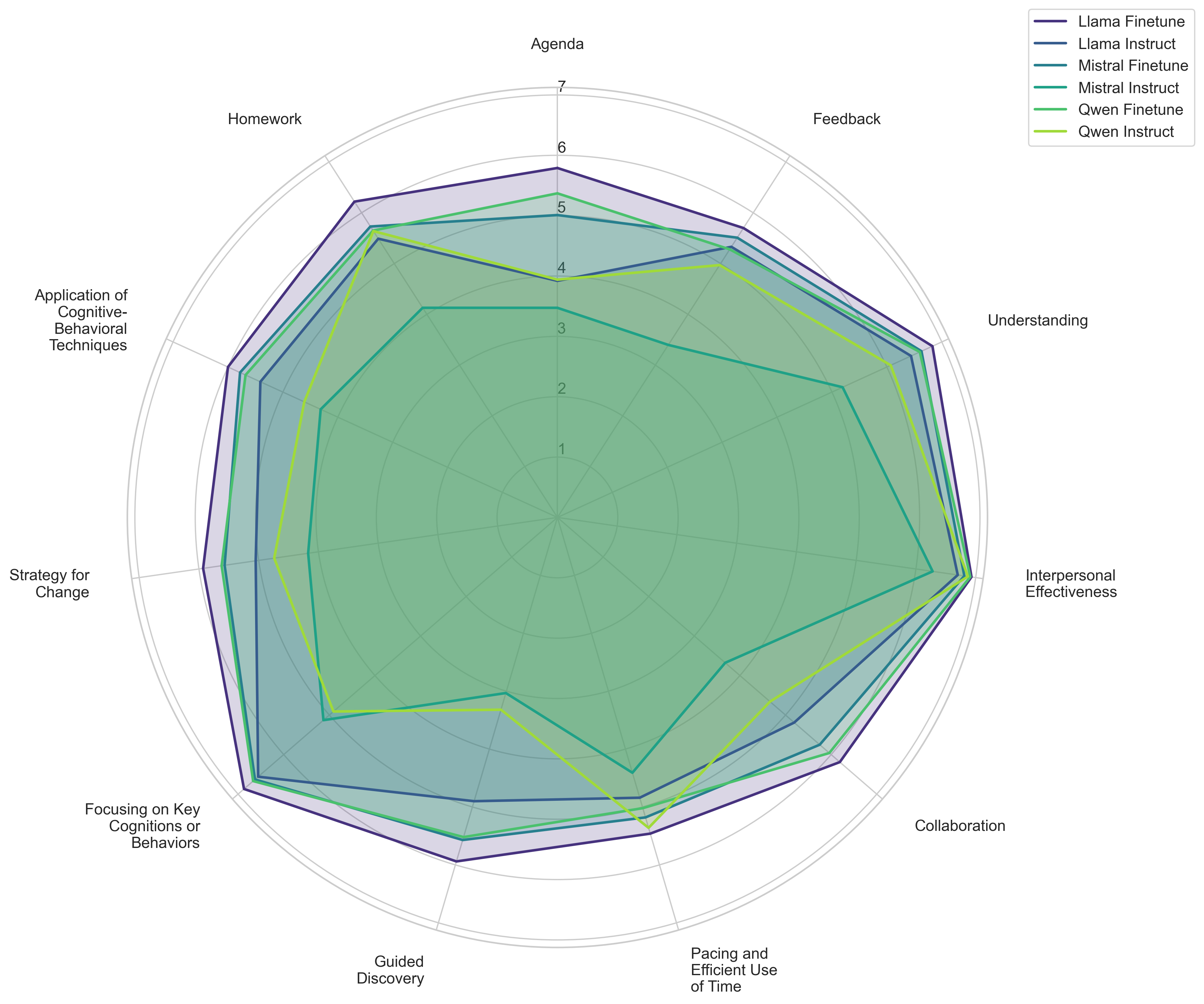}
    \caption{\textbf{Average CTRS category scores by model variant.} The radar chart visualizes the mean performance score for each model variant across the 11 Cognitive Therapy Rating Scale (CTRS) categories. Each axis represents a distinct CTRS category, and the distance from the center indicates the average score. Each colored polygon corresponds to a specific model variant.}
    \label{fig:radar_plot}
\end{figure}

\subsection{Mixed effects model analysis}
A linear mixed effects model was used to analyze CTRS scores, accounting for model type, variant (CBT-tuned vs instruct), and session number, with random intercepts for patient-specific variations. The model configuration and effect estimates are summarized in Table 1. The baseline performance was represented by CBT-tuned Llama 3.1 8b (intercept = 69.51, p < 0.001). The analysis revealed significant differences between models, with Mistral performing 8.05 points lower (p < 0.001) and Qwen performing 4.16 points lower (p < 0.001) compared to Llama. The largest effect was observed when comparing CBT-tuned models to their instruct-tuned counterparts, with instruct models performing 11.33 points lower on average (p < 0.001).

The random effects structure indicated modest variation at the patient level (variance = 0.873). The model found no significant effect of session number ($\beta$ = 0.179, p = 0.675), indicating that performance remained stable over the therapy course. The model was able to explain a significant portion of the variance in CTRS scores, with all main effects showing statistical significance except for session number. 

\begin{table}[h!]
\centering
\begin{threeparttable}
\caption{\textbf{Mixed effects analysis of model performance across variants: Configuration parameters and effect estimates}}
\label{tab:model_analysis}

\begin{subtable}{\linewidth}
\centering
\begin{tabular*}{\linewidth}{@{\extracolsep{\fill}}lclc@{}}
\toprule
\textbf{Model Component} & \textbf{Value} & \textbf{Statistics} & \textbf{Value} \\
\midrule
Model Type & MixedLM & Dependent Variable & total\_score \\
No. Observations & 593 & Method & ML \\
No. Groups & 5 & Scale & 108.3379 \\
Min. group size & 117 & Log-Likelihood & -2232.2854 \\
Max. group size & 120 & Converged & Yes \\
Mean group size & 118.6 & & \\
\bottomrule
\end{tabular*}
\end{subtable}
\vspace{0.5cm}
\begin{subtable}{\linewidth}
\centering
\begin{tabular*}{\linewidth}{@{\extracolsep{\fill}}lrrrrrr@{}}
\toprule
\textbf{Fixed Effects} & \textbf{Coef.} & \textbf{Std.Err.} & \textbf{z} & \textbf{P>|z|} & \textbf{[0.025} & \textbf{0.975]} \\
\midrule
Intercept & 69.508 & 0.948 & 73.307 & 0.000 & 67.650 & 71.367 \\
C(model)[T.Mistral] & -8.054 & 1.050 & -7.667 & 0.000 & -10.113 & -5.995 \\
C(model)[T.Qwen] & -4.160 & 1.041 & -3.997 & 0.000 & -6.200 & -2.120 \\
C(variant)[T.instruct] & -11.326 & 0.855 & -13.246 & 0.000 & -13.002 & -9.650 \\
session\_centered & 0.179 & 0.428 & 0.419 & 0.675 & -0.659 & 1.018 \\
\bottomrule
\end{tabular*}
\end{subtable}
\end{threeparttable}
\end{table}

\subsection{Consistency}
Consistency in model performance was assessed through score distributions and standard deviations. CBT-tuned Llama 3.1 8b was the most consistent model (SD = 7.24) with scores ranging from 47 to 77 points. The most inconsistent was Mistral 7b v0.3, (SD = 14.09) with a CTRS score range of 11 to 76 points. The CBT-tuned versions of all three base models exhibited more stable performance patterns with higher minimum scores and tighter distribution patterns, as seen in Figure 2. \newline

\begin{figure}[h!]
    \centering
    \includegraphics[width=\textwidth]{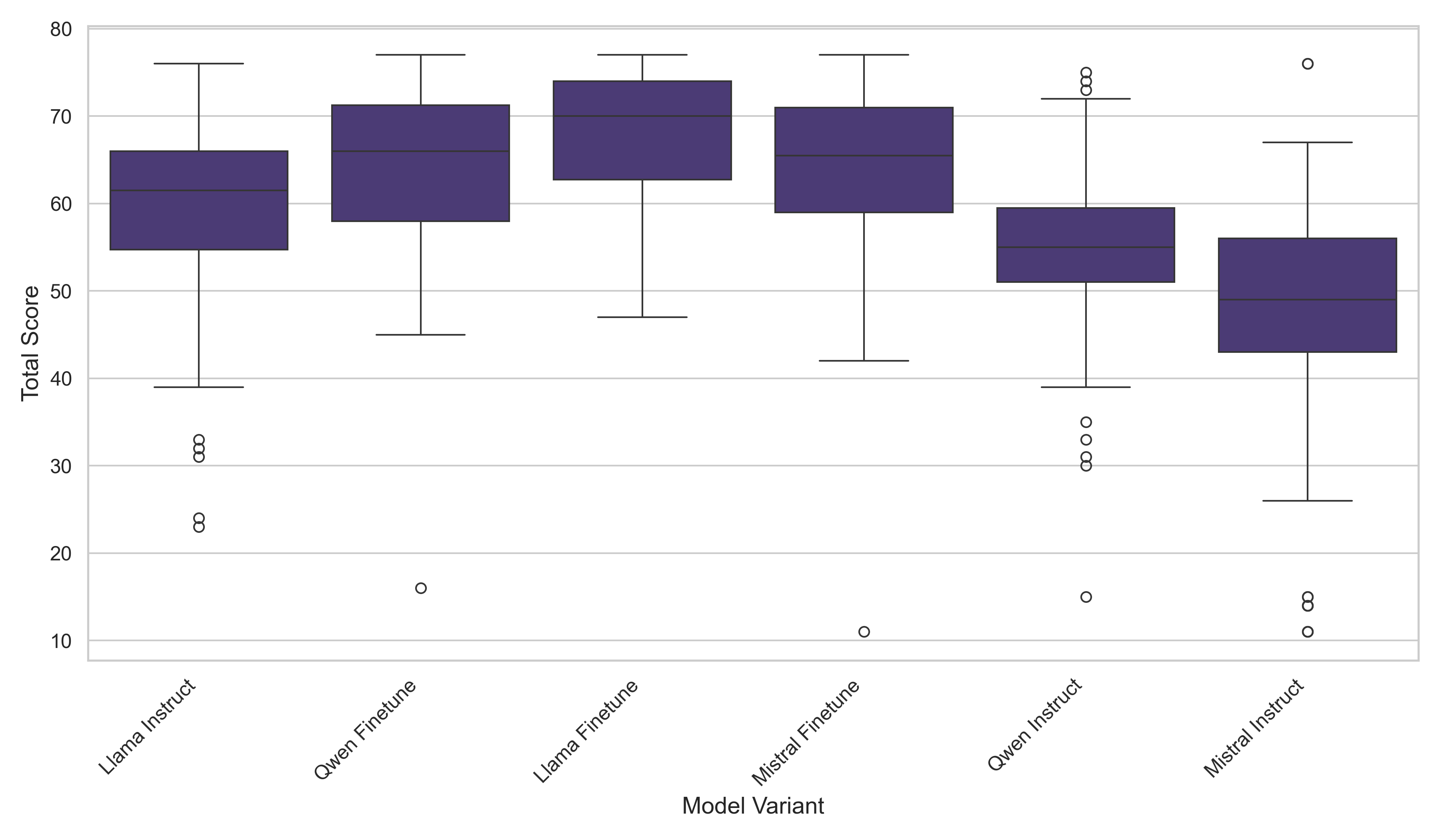}
    \caption{\textbf{Distribution of total CTRS scores by model variant.} The box plot illustrates the distribution of total CTRS scores for each model variant. The central line in each box represents the median score, while the box boundaries indicate the interquartile range (25th to 75th percentiles). Whiskers extend to show the range of the data, and individual points represent outliers.}
    \label{fig:boxplot}
\end{figure}

\subsection{CTRS category analysis}
Analysis of individual CTRS categories revealed distinct patterns of strengths and weaknesses across both model types and variants (CBT-tuned vs instruct). CBT-tuned Llama 3.1 8b was strong across all categories but scored particularly high on Understanding (6.83 ± 0.38), Interpersonal Effectiveness (6.93 ± 0.29), and Focusing on Key Cognitions or Behaviors (6.87 ± 0.49).

The largest differences between CBT-tuned and instruct turned variants were observed in Agenda Setting, Guided Discovery and Application of Cognitive-Behavioral Techniques. In Agenda Setting, CBT-tuned models scored substantially higher (Llama: 5.79 ± 1.72, Mistral: 5.01 ± 1.93, Qwen: 5.37 ± 1.85) compared to their instruct counterparts (Llama: 3.92 ± 1.53, Mistral: 3.47 ± 1.70, Qwen: 3.94 ± 1.75). Guided Discovery also showed one of the greatest disparities, with instruct variants performing particularly poorly (Mistral: 3.03 ± 1.05, Qwen: 3.32 ± 0.99) compared to their CBT-tuned versions (Mistral: 5.57 ± 1.49, Qwen: 5.52 ± 1.49).

All models and variants scored well on Interpersonal Effectiveness (ranging from 6.28 ± 1.28 for Mistral instruct to 6.93 ± 0.29 for Llama CBT-tuned), suggesting that base models already possessed strong capabilities in this domain. Pacing and Efficient Use of Time showed relatively modest scores across all models (ranging from 4.41 ± 1.62 to 5.46 ± 1.29). Meanwhile, the homework category showed the highest variability in scores (SD ranging from 1.56 to 2.41). \newline

\FloatBarrier
\section{Discussion}
Our study demonstrates that fine-tuning open weight LLMs with synthetic CBT transcripts can substantially improve their ability to deliver a full course of CBT for depression. The CBT-tuned variants of the three base models (Llama 3.1 8b, Qwen 2.5 7b, and Mistral 7b v0.3) significantly outperformed their instruct-tuned counterparts as assessed by a modified CTRS. The sizeable difference in CTRS scores between the CBT-tuned and instruct-tuned models (11.33 points on average, p < 0.001), indicates that targeted fine-tuning can effectively encode therapeutic principles and techniques, even in models with a relatively small parameter count. 

The results of our study are particularly encouraging when viewed in the context of the existing literature on AI-delivered CBT. Previous papers have largely relied on prompt engineering with closed-source models and have created systems focused on narrow aspects of CBT delivery. Our work suggests that fine-tuning small open weight LLMs may be a more sustainable, accessible and complete approach to AI-assisted CBT. 

\subsection{Conversation quality and structure}
Qualitative analysis of the simulated patient transcripts and CTRS score justifications provided by Gemini-1.5-Pro-002 reveals several interesting findings. The CBT-tuned models tended to engage in more natural conversation, with shorter responses, a collaborative approach and appropriate turn taking. In contrast, the instruct-tuned models generally produced longer responses, often with numbered or bulleted lists, sometimes overwhelming patients with the amount of information provided. Expectedly, the CBT-tuned models were more consistent in terms of the session structure and integration of key CBT elements, such as agenda setting, check-ins and homework assignment. However, we did observe difficulties with agenda adherence. At times, the check-ins by the CBT-tuned models would run too long or agenda items would be overlooked. These observations align with the quantitative findings of moderate scores in the Pacing and Efficient Use of Time category. It should also be noted that the performance of CBT-tuned models degraded noticeably as the context length approached 4000 tokens, perhaps reflecting the limited long context coherence of the base models and the absence of long context examples in the fine-tuning data. 

\subsection{Implementation of CBT techniques}
The CBT-tuned models demonstrated reasonable competence in socializing the simulated patients to the CBT model and implementing core CBT techniques such as thought records, Socratic questioning and cognitive distortions. Homework assignments were generally appropriate and aligned with SMART goal principles. The CBT-tuned models also consistently reviewed the previous session’s homework at the start of the current session. However, we observed a tendency to introduce cognitive distortions and cognitive restructuring earlier in the therapy course than might be clinically appropriate \cite{Kazantzis2018TheMeta-Analyses, Fenn2013TheTherapy, McGinn2000CognitiveStatus, Wenzel2017BasicTherapy}. The models also failed to engage in a robust exploration of the patient’s depression symptoms and social history in the initial phase of the therapy \cite{McGinn2000CognitiveStatus}. These limitations likely reflect biases in our training data distribution.  One particular strength of the CBT-tuned models, was their consistent acknowledgement of AI limitations at the start of every therapy session. Unfortunately, near the termination of sessions, there were lapses in the models’ abilities to recognize their limitations, as evidenced by offering call and text support to the patient and performing “online” searches for resources. These mistakes suggest significant room for improvement in constraint enforcement.  

\subsection{Therapeutic relationship}
Both the instruct-tuned and CBT-tuned models, scored well in the CTRS categories of Interpersonal Effectiveness and Understanding. All model variants made use of empathetic statements, however, the CBT-tuned models were more likely to engage in regular check-ins and collaborative exploration of patient concerns. Despite, demonstrating effective interpersonal skills, the CBT-tuned models often conducted only a superficial exploration of patient concerns, missing opportunities for deeper examination of thoughts and emotions \cite{Kazantzis2018TheMeta-Analyses, Fenn2013TheTherapy}.

\subsection{Simulation limitations}
There were several limitations we encountered when conducting CBT sessions with simulated patients. We observed that the responses generated by the simulated patients were often unrealistic due to their insightfulness, lack of resistance and desire to follow the therapist’s lead in the session. Simulated patients did not frequently make mistakes or indicate difficulty in understanding CBT techniques, even during first exposures. We were unable to mitigate these issues despite constructing a comprehensive prompt for DeepSeek-V2.5 that encouraged mild resistance, clarification questions and expression of patient preferences. We were also forced to artificially terminate the sessions by checking for termination phrases or using pre-determined word count and turn count limits. All of the aforementioned limitations, potentially increase the simulation-to-reality gap for our fine-tuned models.

\subsection{Coherence}
Although coherence was not quantitatively estimated in our analysis, our observations of the simulated patient transcripts revealed some issues. Both the CBT-tuned and instruct-tuned models demonstrated a tendency to become repetitive over longer contexts and at times, lapsed into role confusion. Role confusion involved the therapist replying as if in the role of the patient or vice versa. Although, role confusion in the CBT-tuned models was rare, there were occasions where the models would produce tokens from the fine-tuning prompt template, such as, ‘\#\#\#’. Furthermore, there was an instance of the CBT-tuned Llama 3.1 8b model referring to a patient by the wrong name and fabricating CBT techniques such as “cognitive maintenance bias”. These errors were not frequent and were sometimes perpetuated the lack of feedback provided by the simulated patient. 

\subsection{Study limitations}
Several important limitations constrain the interpretation of our study findings. Our validation of synthetic transcripts used for fine-tuning relied primarily on visual inspection, and due to the high volume of data, we were unable to thoroughly inspect all transcripts. This may have allowed inconsistencies in our training data to go undetected. Due to compute constraints, we did not conduct hyperparameter optimization which may have led to suboptimal hyperparameter choices and model performance. Compute constraints also prevented us from being able to employ fine-tuning approaches beyond QLoRA, such as LoRA or full fine-tuning, that may have yielded better results. Our synthetic data generation approach, while effective, could be improved by using census demographic data to generate more realistic patient profiles, developing more sophisticated patient resistance patterns in the training data and including a broader range of therapeutic challenges. Furthermore, the fine-tuning data could benefit from examples of crisis situations. Our automated evaluation process also had some significant limitations. By converting the CTRS from a 0-6 to a 1-10 rating scale, using word-based ratings and employing an AI rater, we potentially impacted the validity of the scale. In future studies, these impacts could be mitigated by calibrating the evaluation model with human ratings or fine-tuning it to deliver CTRS ratings at a higher level of agreement with human raters. 

\subsection{Ethics and future directions}
Any work on AI based therapy systems necessitates a discussion of the ethical implications \cite{Mirzaei2024ClinicianImplications, Ma2024NoPsycho-counseling}. Our study was meant to demonstrate the technical feasibility and limitations of creating an AI therapist to deliver CBT. We do not recommend deploying any of the models from our study in clinical applications. This recommendation is made based on the significant limitations of these models including, hallucinations, constraint enforcement and long context coherence. We emphasize that these models should be viewed as tools for further research and as a stepping stone for future efforts to create effective AI therapists rather than clinical aids. 

Derivative studies may want to use higher quality synthetic data with more rigorous validation procedures, larger models and clinical validation. It should be recognized that that therapy, of any modality, has the potential to be harmful for a patient if delivered incorrectly \cite{Castonguay2010TrainingTreatments., Knox2019TheFails}. As such, clinical validation studies are essential to assess real-world applicability. These studies must carefully consider patient safety and include robust monitoring systems for adverse events. As LLMs and AI therapists advance, we will also need to develop clear guidelines for appropriate use cases and contraindications \cite{Stade2024LargeEvaluation, Obradovich2024OpportunitiesPsychiatry}. 

Hallucinations remain a fundamental limitation of LLMs. Furthermore, due to their auto-regressive nature and probabilistic sampling of output tokens, their responses may not always align with expectation or clinical indication \cite{Lappin2024AssessingModels, Xu2024HallucinationModels}. Reinforcement learning with human feedback has been very effective in constraining model responses and providing generally appropriate guardrails for content \cite{Lin2024ReinforcementGiants}. However, in sensitive clinical applications with vulnerable patients, even these guardrails may prove insufficient \cite{Roose2024CanSuicide}.

\section{Conclusion}
This study represents a significant step forward in developing accessible, efficient AI systems capable of delivering CBT-informed interventions. Our results demonstrate that small open-weight language models can be successfully fine-tuned to provide structured therapeutic interactions with reasonable fidelity to CBT principles in simulated environments. The significant improvements observed across all models after CBT fine-tuning suggest that targeted training can effectively encode therapeutic competencies even in models with limited parameters. Future studies should focus on addressing the identified technical limitations, enhancing quality of training data, refining evaluation methodologies, and clinical validation.

\section*{Conflicts of Interest} 
The author declares no conflicts of interest that could have influenced the conduct, analysis, or reporting of this research.

\section*{Funding}
This research was independently funded by the primary author (T.T.) without external financial support from any agency, institution, or organization. 

\section*{Data availability}
The data, codebase, and implementation details for this study are publicly available under MIT License in the project's \href{https://github.com/ttahir-git/FineTuning_LLMs_for_CBT_for_Depression.git}{\textit{GitHub Repository}}.
All fine-tuned models developed in this study are publicly available through the Hugging Face model repository - \href{https://huggingface.co/TTahir}{\textit{HuggingFace/TTahir}}.

\bibliography{references} 

\end{document}